\documentclass[]{article}
\usepackage[a4paper]{geometry}
\usepackage{mtsummit2023}
\usepackage{times}
\usepackage{url}
\usepackage{latexsym}
\usepackage{natbib}
\usepackage{layout}
\usepackage{graphicx}
\usepackage[hang,marginal]{footmisc}
\usepackage{multirow}
\usepackage{booktabs, multirow} % for borders and merged ranges
\usepackage{soul}% for underlines
\usepackage[table]{xcolor} % for cell colors
\usepackage{changepage,threeparttable} % for wide tables

\usepackage{array,color}
\newcolumntype{R}{>{$\mathcode`\-"8000\relax\color{blue}}r<$}
\makeatletter
{\catcode`\-\active
\gdef-{\color{red}\mathchar"2200\relax}}
\makeatletter
\begin{document}

% \amtaHeader{x}{x}{xxx-xxx}{2015}{45-character paper description goes here}{Author(s) initials and last name go here}
\title{\bf Character-level NMT and language similarity}

\author{\name{\bf Josef Jon } \hfill  \addr{jon@ufal.mff.cuni.cz}\\
\name{\bf  Ondřej Bojar} \hfill  \addr{bojar@ufal.mff.cuni.cz} \\ 
\addr{Charles University, Faculty of Mathematics and Physics, Institute of Formal and Applied Linguistics, Prague, Czech Republic}
}

\maketitle
\pagestyle{empty}

\begin{abstract}

  We explore the effectiveness of character-level neural machine translation using Transformer architecture for various levels of language similarity and size of the training dataset on translation between Czech and Croatian, German, Hungarian, Slovak, and Spanish. We evaluate the models using automatic MT metrics and show that translation between similar languages benefits from character-level input segmentation, while for less related languages, character-level vanilla Transformer-base often lags behind subword-level segmentation. We confirm previous findings that it is possible to close the gap by finetuning the already trained subword-level models to character-level.
\end{abstract}
\section{Introduction}
Character-level NMT has been studied for a long time, with mixed results compared to subword segmentation. In the MT practitioner's discourse, it has sometimes been assumed that character-level systems are more robust to domain shift and better in the translation of morphologically rich languages. Recent studies \citep{libovicky-etal-2022-dont} show that there are no conclusive proofs for these claims.

At the same time, character-level systems have been reliably shown to be robust against source-side noise. In terms of general translation quality, they often either underperform or are on par with their subword-level counterparts \citep{libovicky-etal-2022-dont}.  Also, both training and inference speeds are lower and memory requirements are higher due to longer sequence lengths (mostly because of the quadratic complexity of the Transformer attention mechanism with respect to the input length \citep{vaswani2017attention}) unless specialized architectures are used.

In this work, we present experiments on a specific use-case of translation of related languages. We train bilingual Transformer translation models to translate between Czech and Croatian, German, Hungarian, Slovak, or Spanish. We vary the training dataset size, vocabulary size and model depth and study the effects.  We show that in the baseline configuration with vanilla \texttt{Transformer-base}, character-level models outperform subword-level models in terms of automated evaluation scores only in closely related Czech-Slovak translation pair. Finally, we confirm that it is possible to obtain a better quality of the char-level translation for less related languages by first training a subword-level model and in the later stage of the training switching to character-level processing.

\section{Related work}
\citet{libovicky-etal-2022-dont} analyze the body of the work on character-level NMT and show that in most cases, it still falls behind in many aspects compared to the subword-level counterpart. Since they provide a comprehensive overview of the field up to today, we will only very briefly list the most influential works in this section, and refer the reader to the detailed analysis in \citet{libovicky-etal-2022-dont}.

In one of the earliest works, \citet{chung-etal-2016-character} use RNN with character segmentation on the decoder side. \citet{lee-etal-2017-fully} use CNN for fully character-level NMT. \citet{costa-jussa-etal-2017-byte} apply a similar approach to byte-level translation.
\citet{gupta2019characterbased} and \citet{ngo-etal-2019-Transformer} explore character-level MT using the Transformer model.
Recent work on character-level NMT includes \citet{li-etal-2021-char,banar} and \citet{gao-etal-2020-character}.

\citet{libovicky-fraser-2020-towards} show that problems with slow training and worse final translation quality for character-level NMT models can be largely mitigated by first training with subword segmentation and subsequently finetuning on character-segmented text. However, a problem of lower speed (due to longer sequence length) persists, which can make both the training and inference prohibitively costly and slow, especially for models that make use of a larger context than only one sentence.

Our work specifically targets character-level translation of closely related languages. In WMT 2019 Similar Language translation task \citep{barrault-etal-2019-findings}, \citet{scherrer-etal-2019-university} show that  character-level NMT is effective for translation between closely related Portuguese and Spanish and in Multilingual Low-Resource Translation for Indo-European Languages task at WMT21 \citep{akhbardeh-etal-2021-findings}, \cite{jon-etal-2021-cuni} successfully apply character-level NMT to translation between Catalan and Occitan. 

\section{System description}
\subsection{Data}
We evaluate our models on translation from Czech to German, Spanish, Croatian, Hungarian and Slovak and vice-versa.
We train on MultiParaCrawl \citep{banon-etal-2020-paracrawl}\footnote{https://opus.nlpl.eu/MultiParaCrawl.php} corpus. It is based on Paracrawl, which is English-centric (each language in the original dataset is aligned only to English). MultiParaCrawl aligns the sentences in the other languages that have the same English translation. This introduces mis-alignments into the dataset (it is possible that two sentences with different meanings in other languages have the same English translation), but we nevertheless use it to have a comparable training corpus for all the languages. We sample subsets for each language pair in sizes of 50k, 500k, and 5M sentences (Croatian corpus only has about 800k sentences in total, so we use only the 50k and 500k sizes). 
We use FLORES-200 \citep{nllb2022} as validation and test sets (we keep the original splits). We note that this test set is created by translating the same English test into all the languages and not translating the two tested languages between each other -- this might mean that the effect of language similarity is somewhat subdued in this setting.

We segment the text using SentencePiece with the given vocabulary size (32k, 4k, or character-level model), with 99.95\% character coverage and UTF-8 byte fallback for unknown characters. The segmentation models are trained on the whole 5M datasets, jointly for each pair. 
\paragraph{Language similarity} 
We use chrF score \citep{popovic-2015-chrf}, traditionally used to compute translation quality, as a language similarity metric. It is a character-level metric and we hypothesize that character-level similarity is an important aspect for our experiments. We compute chrF score of the Czech FLORES-200 test set relative to all the other languages (Table \ref{tab:sim}). We also show the lexical similarity score provided by the UKC database\footnote{http://ukc.disi.unitn.it/index.php/lexsim/}, which is based on a number of cognates between languages in their contemporary vocabularies \citep{bella}.
\begin{table}[]
\centering 
\parbox{.45\linewidth}{
\centering
\begin{tabular}{lrr}
\textbf{Language} & \multicolumn{1}{l}{\textbf{chrF}} & \textbf{LexSim} \\ \hline

\textbf{sk}       & 36.7     &     16.5                    \\
\textbf{hr}       & 22.7         &     8.2                \\
\textbf{es}       & 16.5        &            2.6          \\
\textbf{hu}       & 16.3      &                 2.9      \\
\textbf{de}       & 15.4       &          3.7             \\
\end{tabular}
\caption{UKC LexSim and chrF score-based similarities of the testsets, i.e. chrF score of untranslated Czech testset compared to the other languages.} %\XXX{I should have a look at this similarity for camer ready or next submission http://ukc.disi.unitn.it/index.php/lexsim/}}
\label{tab:sim}

}
\hspace{15px}
\parbox{.45\linewidth}{
\centering
\begin{tabular}{ccrr}
\textbf{Pair}     & \multicolumn{1}{c}{\textbf{Lang}} & \multicolumn{1}{c}{\textbf{\% skip}} & \multicolumn{1}{c}{\textbf{Avg len}} \\ \hline
\multirow{2}{*}{cs-de} & cs                                & 0.43                                    & 88.2                                 \\
                       & de                                & 0.64                                    & 100.3                                \\
\multirow{2}{*}{cs-es} & cs                                & 0.30                                    & 84.5                                 \\
                       & es                                & 0.50                                    & 95.5                                 \\
\multirow{2}{*}{cs-hr} & cs                                & 1.21                                    & 127.1                                \\
                       & hr                                & 1.30                                    & 131.7                                \\
\multirow{2}{*}{cs-hu} & cs                                & 0.26                                    & 76.4                                 \\
                       & hu                                & 0.45                                    & 83.0                                 \\
\multirow{2}{*}{cs-sk} & cs                                & 0.25                                    & 74.9                                 \\
                       & sk                                & 0.29                                    & 77.4                                
\end{tabular}
\caption{Percentage of examples exceeding the training source length limit (400 characters) and average sentence character lengths for all the training datasets for character-level training. }%\XXX{Keep this? It's here because Table 2 looks weird alone in one-column format}}
\label{tab:lengths}

}

\end{table}

\subsection{Model}
\label{sec:model}
We trained Transformer \citep{vaswani2017attention} models to translate to Czech from other languages (Hungarian, Slovak, Croatian, German and Spanish) and vice-versa using MarianNMT \citep{junczys-dowmunt-etal-2018-marian}.

Our baseline model is \texttt{Transformer-base} (512-dim embeddings, 2048-dim ffn) with 6 encoder and 6 decoder layers. We also train two other versions of \texttt{Transformer-base}: with 16 encoder + 6 decoder layers and %, \texttt{Transformer-base} with 16 encoder and 16 decoder layers and \texttt{transfomer-big} (1024-dim embeddings, 4096-dim ffn) with 16 encoder and 16 decoder layers. 
with 16 encoder + 16 decoder layers. For other hyper-parameters, we use the default configuration of MarianNMT.
We evaluate the models on the validation set each 5000 updates and we stop the training after 20 consecutive validations without improvement in either chrF or cross-entropy. We use Adam optimizer~\citep{kingma2017adam} and one shared vocabulary and embeddings for both source and target.

Similarly to \citet{libovicky-fraser-2020-towards}, we compared training char-level models from scratch to starting the training  from subword-level models (both with 4k and 32k vocabularies) and switching to character-level processing after subword-level training converged. They obtained better results with a more complex curriculum learning scheme, while we only finetune the pre-trained model.

We performed a length analysis on the character level for all the datasets. Based on this, we set the maximum source sequence length for training and inference to 400 for all the systems. We skip longer training examples. In the worst case (Croatian to Czech), 1.3 \% of the examples are skipped. Table \ref{tab:lengths} shows average character lengths and percentage of the skipped training examples in all directions. For inference, we normalize the output score by the length of the hypothesis as implemented in Marian. We search for the optimal value of the length normalization constant on the validation set in the range of 0.5 to 4.0.

\subsection{Evaluation}

  We use SacreBLEU~\citep{post-2018-call} to compute BLEU and chrF scores. We set $\beta=2$ for chrF in all the experiments (i.e. chrF2, the default in SacreBLEU). For COMET~\citep{rei-etal-2020-comet}\footnote{\url{https://github.com/Unbabel/COMET}} scores we use the original implementation and the \texttt{wmt20-comet-da} model.

\subsection{Hardware}
We ran the experiments on a grid comprising of Quadro RTX 5000, GeForce GTX 1080 Ti, RTX A4000, or GeForce RTX 3090 GPUs. We trained a total of about 170 models with training times ranging from 10 hours to 14 days, depending on the dataset, model, and GPUs used. 

\section{Results}

\subsection{Subwords vs. characters}
% \usepackage{multirow}% Please add the following required packages to your document preamble:
% \usepackage{multirow}
\begin{table}[!htp]\centering
\small
\begin{tabular}{lrr@{\hskip 30px}rrr@{\hskip 30px}rrrr}
& & &\multicolumn{3}{c}{\textbf{Czech $\rightarrow$ Lang}} &\multicolumn{3}{c}{\textbf{Lang $\rightarrow$ Czech}} \\\cmidrule{4-9}
\textbf{Lang} &\textbf{Dataset} &\textbf{Vocab} &\textbf{BLEU} &\textbf{CHRF} &\textbf{COMET} &\textbf{BLEU} &\textbf{CHRF} &\textbf{COMET} \\\midrule
\multirow{9}{*}{sk} &\multirow{3}{*}{50k} &char &\textbf{23.1} &\textbf{53.1} &\textbf{0.8834} &\textbf{23.4} &\textbf{53.1} &\textbf{0.8429} \\
& &4k &21.1 &51.7 &0.6989 &21.6 &51.8 &0.7054 \\
& &32k &20.1 &50.5 &0.5155 &20.1 &50.2 &0.5226 \\ \cmidrule{2-9}
&\multirow{3}{*}{500k} &char &\textbf{27.8} &\textbf{56.4} &\textbf{1.0737} &\textbf{27.2} &\textbf{56.1} &\textbf{1.0165} \\
& &4k &27.0 &55.8 &1.0574 &26.7 &55.8 &1.0018 \\
& &32k &26.8 &55.6 &1.0342 &26.3 &55.4 &0.9893 \\  \cmidrule{2-9}
&\multirow{3}{*}{5M} &char &\textbf{28.7} &\textbf{57.0} &\textbf{1.1035} &\textbf{28.4} &\textbf{56.8} &\textbf{1.0419} \\
& &4k &28.6 &56.9 &1.1012 &28.1 &56.5 &1.0333 \\
& &32k &\textbf{28.7} &56.9 &1.0973 &28.2 &56.6 &1.0376 \\ \midrule
\multirow{9}{*}{hu} &\multirow{3}{*}{50k} &char &0.6 &21.0 &-1.4054 &0.3 &18.1 &-1.4137 \\
& &4k &1.9 &25.4 &-1.3256 &1.5 &24.2 &-1.2826 \\
& &32k &\textbf{3.0} &\textbf{28.3} &\textbf{-1.2141} &\textbf{2.1} &\textbf{25.5} &\textbf{-1.2116} \\   \cmidrule{2-9}
&\multirow{3}{*}{500k} &char &\textbf{13.3} &\textbf{45.8} &\textbf{0.1812} &\textbf{12.3} &\textbf{42.2} &0.1892 \\
& &4k &12.7 &44.7 &0.1371 &\textbf{12.3} &41.2 &\textbf{0.2414} \\
& &32k &12.4 &43.4 &0.0852 &11.8 &40.6 &0.1658 \\  \cmidrule{2-9}
&\multirow{3}{*}{5M} &char &17.4 &\textbf{50.8} &0.6263 &17.7 &46.9 &0.6999 \\
& &4k &17.7 &50.3 &0.6447 &18.4 &\textbf{47.4} &0.7283 \\
& &32k &\textbf{18.3} &50.6 &\textbf{0.6531} &\textbf{18.6} &47.2 &\textbf{0.7325} \\ \midrule
\multirow{9}{*}{de} &\multirow{3}{*}{50k} &char &0.4 &22.5 &-1.5904 &0.4 &18.5 &-1.4006 \\
& &4k &2.2 &29.2 &-1.3982 &2.0 &25.7 &-1.2548 \\
& &32k &\textbf{4.7} &\textbf{33.7} &\textbf{-1.2014} &\textbf{4.7} &\textbf{29.9} &\textbf{-1.0102} \\  \cmidrule{2-9}
&\multirow{3}{*}{500k} &char &18.0 &50.6 &0.3185 &\textbf{18.0} &\textbf{47.3} &0.4657 \\
& &4k &\textbf{19.2} &\textbf{50.9} &\textbf{0.3568} &\textbf{18.0} &\textbf{47.3} &\textbf{0.5533} \\
& &32k &\textbf{19.2} &50.3 &0.3155 &17.6 &46.1 &0.4517 \\  \cmidrule{2-9}
&\multirow{3}{*}{5M} &char &24.1 &55.2 &0.5955 &23.1 &\textbf{52.0} &0.8322 \\
& &4k &24.3 &55.2 &0.6043 &23.0 &51.9 &0.8648 \\
& &32k &\textbf{25.2} &\textbf{55.7} &\textbf{0.6275} &\textbf{23.4} &51.8 &\textbf{0.8838} \\ \midrule
\multirow{9}{*}{es} &\multirow{3}{*}{50k} &char &0.2 &23.0 &-1.4847 &0.2 &18.3 &-1.3952 \\
& &4k &2.3 &28.4 &-1.329 &1.4 &24.0 &-1.2688 \\
& &32k &\textbf{4.6} &\textbf{32.6} &\textbf{-1.1684} &\textbf{2.8} &\textbf{27.3} &\textbf{-1.0927} \\  \cmidrule{2-9}
&\multirow{3}{*}{500k} &char &16.0 &\textbf{46.6} &\textbf{0.1857} &0.4 &18.1 &-1.3986 \\
& &4k &15.6 &45.7 &0.1765 &\textbf{11.7} &\textbf{41.2} &\textbf{0.3451} \\
& &32k &\textbf{15.8} &45.4 &0.0976 &11.5 &40.2 &0.2395 \\ \cmidrule{2-9}
&\multirow{3}{*}{5M} &char &19.3 &\textbf{49.5} &0.4602 &14.6 &44.2 &0.6394 \\  
& &4k &20.0 &49.3 &0.4911 &\textbf{15.7} &44.9 &0.7160 \\
& &32k &\textbf{20.4} &49.4 &\textbf{0.5074} &\textbf{15.7} &\textbf{45.1} &\textbf{0.7186} \\ \midrule
\multirow{6}{*}{hr} &\multirow{3}{*}{50k} &char &0.2 &21.2 &-1.4055 &0.2 &16.9 &-1.4397 \\
& &4k &4.8 &34.0 &-1.0112 &4.6 &30.3 &-1.0283 \\
& &32k &\textbf{7.7} &\textbf{38.1} &\textbf{-0.7048 }&\textbf{5.3 }&\textbf{31.3 }&\textbf{-0.9501} \\  \cmidrule{2-9}
&\multirow{3}{*}{500k} &char &19.6 &\textbf{51.6} & 0.6403 &  18.0 & 47.3 & 0.5469 \\
& &4k &\textbf{19.7} &51.2 &\textbf{0.6922} &\textbf{19.3} &\textbf{48.2} &\textbf{0.6772} \\
& &32k &19.2 &50.5 &0.6160 &19.3 &47.6 &0.6170 \\
\bottomrule
\end{tabular}
\caption{Test set scores for Transformer-base models (6 encoder and 6 decoder layers) trained from scratch. Bold are the best results within the same training dataset.}\label{tab:baseline}

\end{table}

\begin{figure}[htp]
\vspace{-40px}
\includegraphics[width=.34\textwidth]{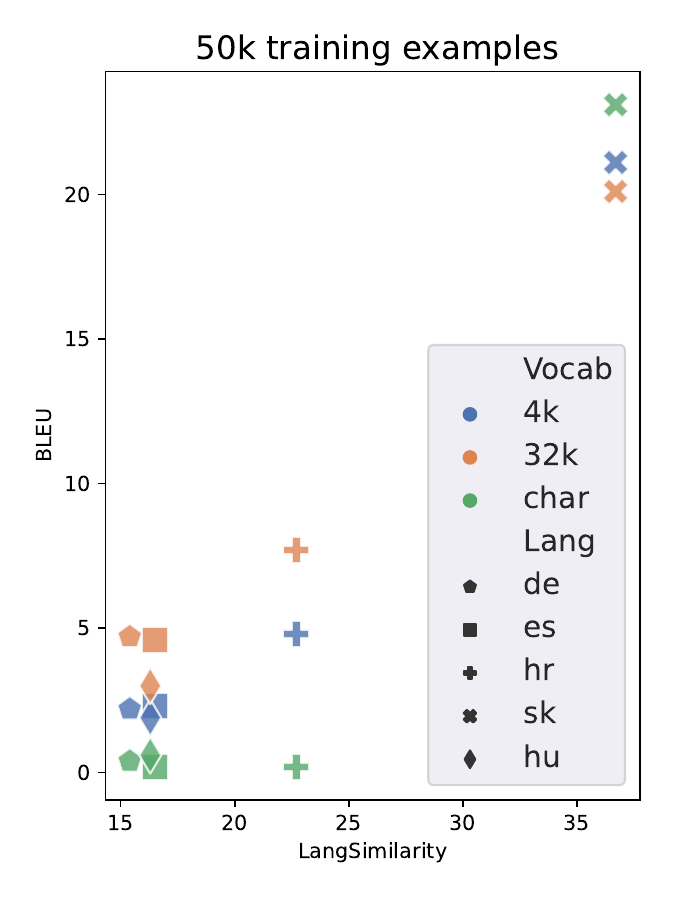}\hfill\hspace{-10px}
\includegraphics[width=.34\textwidth]{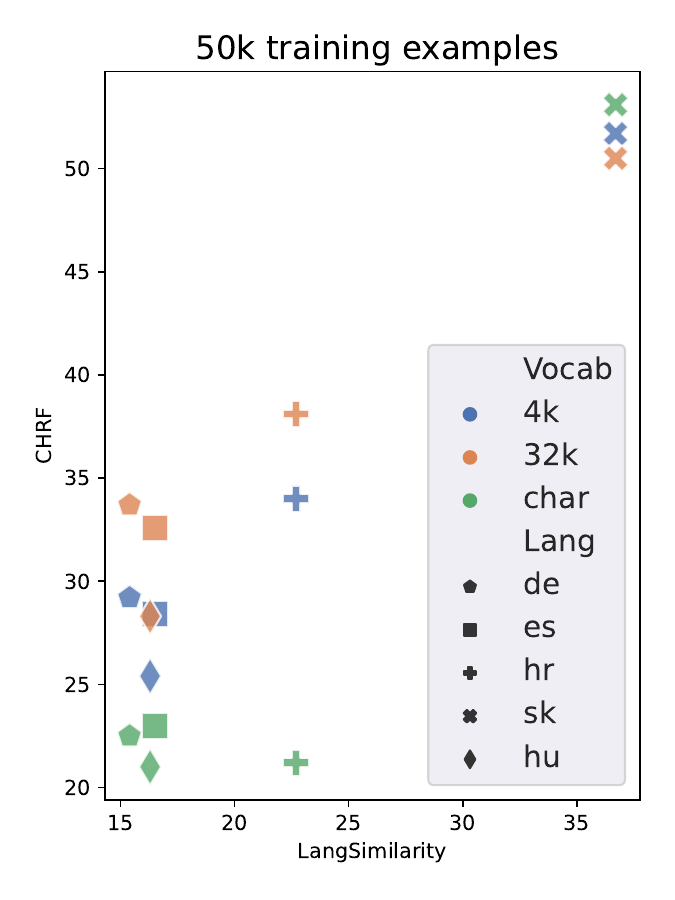}\hfill\hspace{-10px}
\includegraphics[width=.34\textwidth]{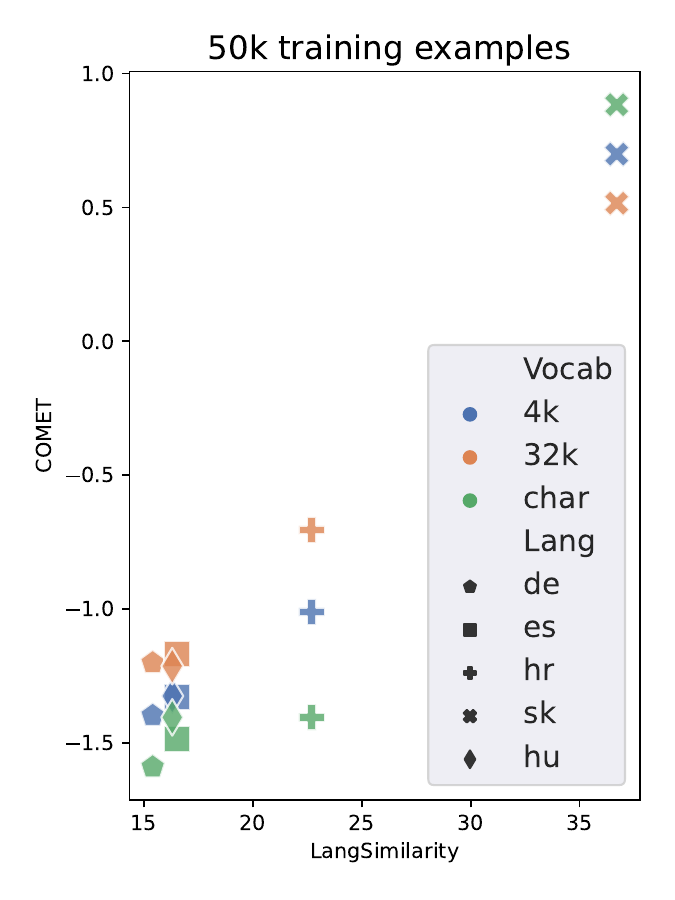}

\includegraphics[width=.34\textwidth]{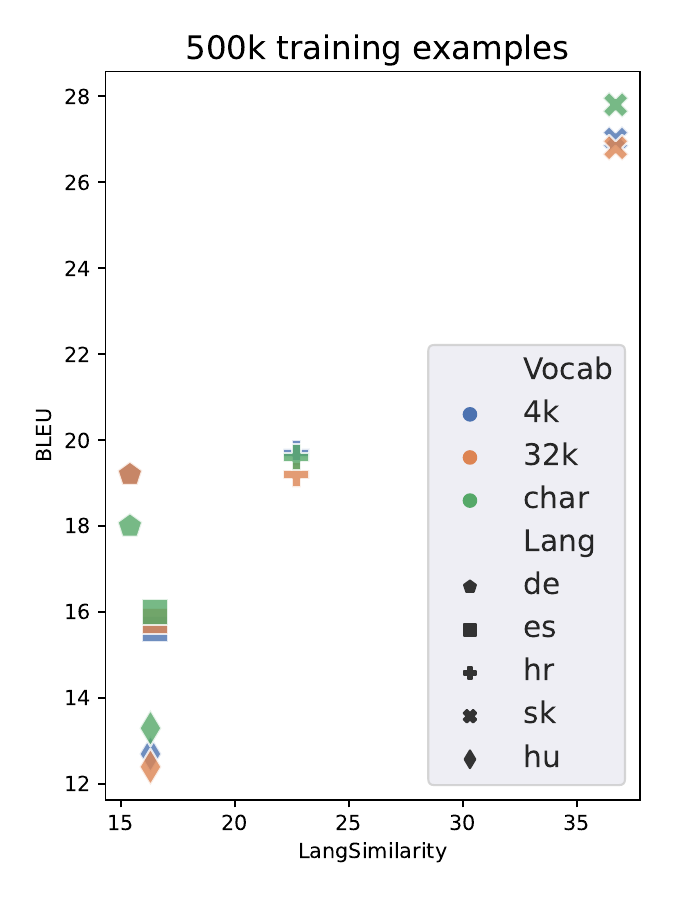}\hfill\hspace{-10px}
\includegraphics[width=.34\textwidth]{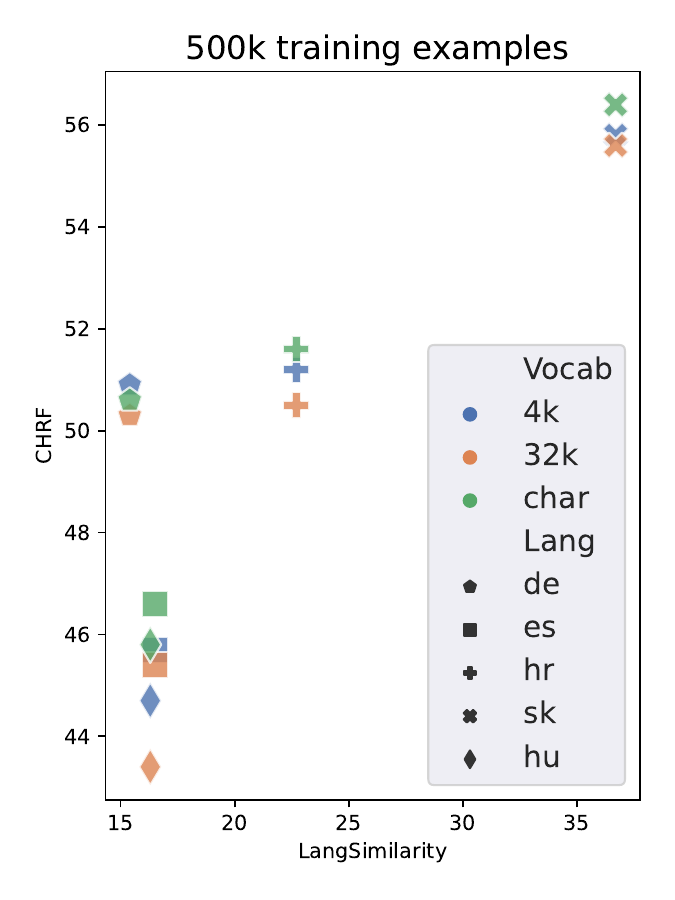}\hfill\hspace{-10px}
\includegraphics[width=.34\textwidth]{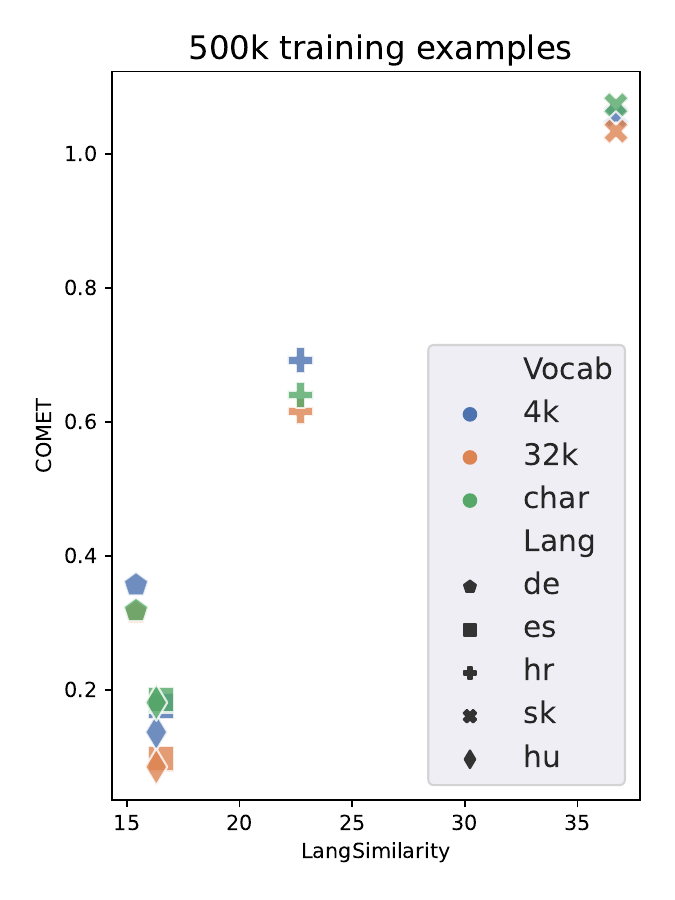}

\includegraphics[width=.34\textwidth]{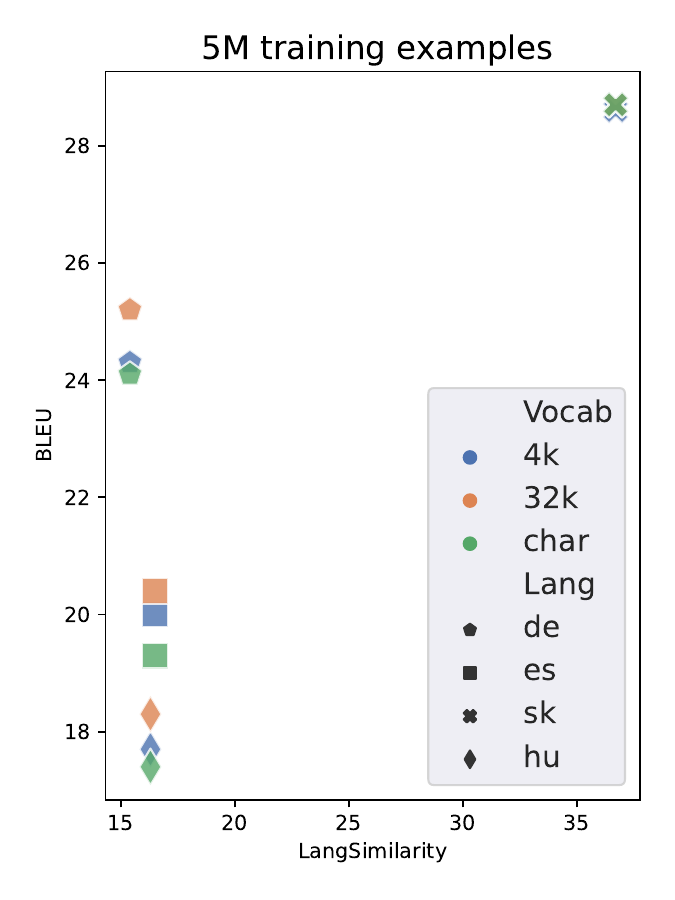}\hfill\hspace{-10px}
\includegraphics[width=.34\textwidth]{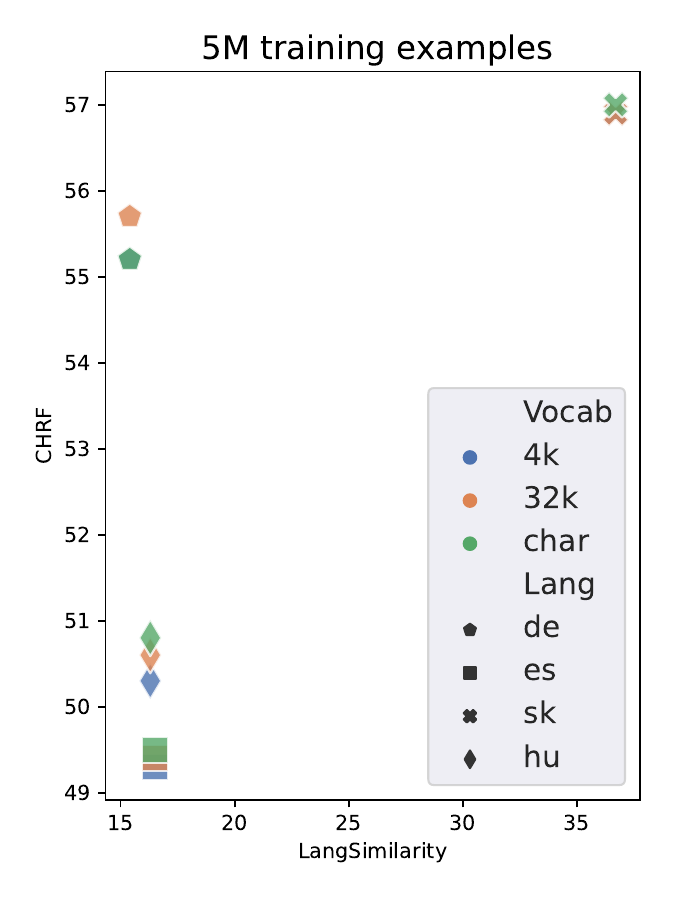}\hfill\hspace{-10px}
\includegraphics[width=.34\textwidth]{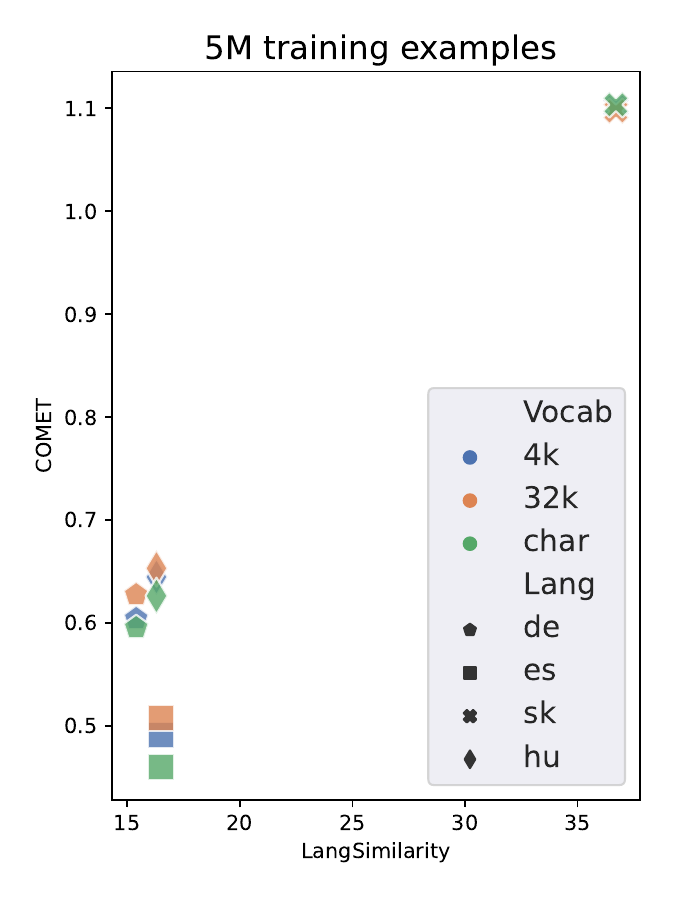}
\caption{Relationship between language similarity scores (chrF of the untranslated test set source) and BLEU, chrF and COMET scores, depending on vocabulary size. First row are the results for 50k sentence train set, second row for 500k train set and third row for 5M train set.}
\label{fig:sim_vs_vocab}

\end{figure}

\begin{table}[!htp] \centering
\scriptsize
\begin{tabular}{lrrrrRRRRRR}
 & &Score & & & \multicolumn{3}{c}{$\Delta(char)$}& \multicolumn{3}{c}{$\Delta(4k)$} \\  \cmidrule{3-11}
Lang &Dataset &BLEU &CHRF &COMET & \multicolumn{1}{r}{BLEU} &\multicolumn{1}{r}{CHRF} &\multicolumn{1}{r}{COMET} &\multicolumn{1}{r}{BLEU} &\multicolumn{1}{r}{CHRF} &\multicolumn{1}{r}{COMET} \\\midrule

sk &50k &21.8 &52.4 &0.8750 &-1.3 &-0.7 &-0.0084 &0.7 &0.7 &0.1761 \\
&500k &27.6 &56.3 &1.0720 &-0.2 &-0.1 &-0.0017 &0.6 &0.5 &0.0146 \\
&5M &28.8 &57.0 &1.1017 &0.1 &0.0 &-0.0018 &0.2 &0.1 &0.0005 \\ \midrule
hu &50k &1.7 &22.8 &-1.3850 &1.1 &1.8 &0.0204 &-0.2 &-2.6 &-0.0594 \\
&500k &13.4 &46.0 &0.2555 &0.1 &0.2 &0.0743 &0.7 &1.3 &0.1184 \\
&5M &18.2 &51.2 &0.6726 &0.8 &0.4 &0.0463 &0.5 &0.9 &0.0279 \\ \midrule
de &50k &2.9 &30.7 &-1.4227 &2.5 &8.2 &0.1677 &0.7 &1.5 &-0.0245 \\
&500k &19.3 &51.3 &0.3966 &1.3 &0.7 &0.0781 &0.1 &0.4 &0.0398 \\
&5M &24.7 &55.6 &0.6214 &0.6 &0.4 &0.0259 &0.4 &0.4 &0.0171 \\ \midrule
es &50k &1.8 &27.5 &-1.4024 &1.6 &4.5 &0.0823 &-0.5 &-0.9 &-0.0734 \\
&500k &16.3 &46.4 &0.2276 &0.3 &-0.2 &0.0419 &0.7 &0.7 &0.0511 \\
&5M &19.8 &49.5 &0.5038 &0.5 &0.0 &0.0436 &-0.2 &0.2 &0.0127 \\ \midrule
hr &50k &10.3 &42.9 &-0.2671 &10.1 &21.7 &1.1384 &5.5 &8.9 &0.7441 \\
&500k &20.6 &52.4 &0.7382 &1.0 &0.8 &0.0979 &0.9 &1.2 &0.0460 \\
\bottomrule
\end{tabular}
\caption{Results of char-level models for translation from Czech finetuned from 4k subword-level models. Numbers under $\Delta(char)$ show the difference between fine-tuned model scores compared to the char-level model trained from scratch, under $\Delta(4k)$ difference from the model that served as the initial checkpoint for the finetuning.} \label{tab:finetuned4}

\end{table}

\begin{table}[!htp] \centering
\scriptsize
\begin{tabular}{lrrrrRRRRRR}
 & &Score & & & \multicolumn{3}{c}{$\Delta(char)$}& \multicolumn{3}{c}{$\Delta(32k)$} \\  \cmidrule{3-11}
Lang &Dataset &BLEU &CHRF &COMET & \multicolumn{1}{r}{BLEU} &\multicolumn{1}{r}{CHRF} &\multicolumn{1}{r}{COMET} &\multicolumn{1}{r}{BLEU} &\multicolumn{1}{r}{CHRF} &\multicolumn{1}{r}{COMET} \\\midrule

\multirow{3}{*}{sk}  &50k &21.2 &52.2 &0.8697 &-1.9 &-0.9 &-0.0137 &1.1 &1.7 &0.3542 \\
&500k &27.5 &56.2 &1.0723 &-0.3 &-0.2 &-0.0014 &0.7 &0.6 &0.0381 \\
&5M &29 &57.2 &1.1011 &0.3 &0.2 &-0.0024 &0.3 &0.3 &0.0038 \\ \midrule
\multirow{3}{*}{hu}  &50k &2.2 &24.8 &-1.358 &1.6 &3.8 &0.0474 &-0.8 &-3.5 &-0.1439 \\
&500k &12.7 &45.7 &0.1832 &-0.6 &-0.1 &0.0020 &0.3 &2.3 &0.0980 \\
&5M &18 &51.0 &0.6589 &0.6 &0.2 &0.0326 &-0.3 &0.4 &0.0058 \\ \midrule
\multirow{3}{*}{de}  &50k &4.5 &33.3 &-1.3335 &4.1 &10.8 &0.2569 &-0.2 &-0.4 &-0.1321 \\
&500k &19.4 &51.4 &0.3775 &1.4 &0.8 &0.0590 &1.4 &0.8 &0.0590 \\
&5M &24.8 &55.6 &0.6274 &0.7 &0.4 &0.0319 &-0.4 &-0.1 &-0.0001 \\ \midrule
\multirow{3}{*}{es}   &50k &3.3 &30.9 &-1.3182 &3.1 &7.9 &0.1665 &-1.3 &-1.7 &-0.1498 \\
&500k &15.8 &46.2 &0.1854 &-0.2 &-0.4 &-0.0003 &0.0 &0.8 &0.0878 \\
&5M &19.6 &49.4 &0.4875 &0.3 &-0.1 &0.0273 &-0.8 &0.0 &-0.0199 \\ \midrule
\multirow{2}{*}{hr}  &50k &8.9 &41.3 &-0.4144 &8.7 &20.1 &0.9911 &1.2 &3.2 &0.2904 \\
&500k &20.5 &52.0 &0.7181 &0.9 &0.4 &0.0778 &1.3 &1.5 &0.1021 \\
\bottomrule
\end{tabular}
\caption{Results of char-level models for translation from Czech finetuned from 32k subword-level models. Numbers under $\Delta(char)$ show the difference between fine-tuned model scores compared to the char-level model trained from scratch, under $\Delta(32k)$ difference from the model that served as the initial checkpoint for the finetuning.} \label{tab:finetuned32}

\end{table}

\begin{table}[!htp]\centering
\footnotesize

\begin{tabular}{lrr@{\hskip 25px}rrr@{\hskip 25px}rrr}
 & & &\multicolumn{3}{c}{\textbf{16-enc/6-dec}} &\multicolumn{3}{c}{\textbf{16-enc/16-dec}} \\ \cmidrule{4-6} \cmidrule{6-9}
\textbf{Lang} &\textbf{Dataset} &\textbf{Vocab} &\textbf{BLEU} &\textbf{CHRF} &\textbf{COMET} &\textbf{BLEU} &\textbf{CHRF} &\textbf{COMET} \\\midrule
\multirow{9}{*}{sk} &\multirow{3}{*}{50k} &char &\textbf{21.9} &\textbf{52.4} &\textbf{0.8475} &\textbf{21.9} &\textbf{52.0} &\textbf{0.8001} \\
& &4k &20.2 &51.0 &0.6444 &19.3 &50.1 &0.5262 \\
& &32k &19.6 &50.1 &0.5308 &20.1 &50.4 &0.5764 \\  \cmidrule{2-9}
&\multirow{3}{*}{500k} &char &\textbf{27.4} &\textbf{56.0} &\textbf{1.0621} &\textbf{27.4} &\textbf{56.1} &\textbf{1.0618} \\
& &4k &26.5 &55.6 &1.0432 &26.6 &55.6 &1.0469 \\
& &32k &26.2 &55.4 &1.0319 &26.2 &55.4 &1.0194 \\  \cmidrule{2-9}
&\multirow{3}{*}{5M} &char &\textbf{28.6} &\textbf{57.0} &\textbf{1.1016} &\textbf{28.5} &\textbf{56.9} &\textbf{1.1013} \\
& &4k &\textbf{28.6} &56.9 &1.1015 &28.3 &56.7 &1.0920 \\
& &32k &28.2 &56.7 &1.0916 &28.4 &56.8 &1.0986 \\  \midrule
\multirow{9}{*}{hu} &\multirow{3}{*}{50k} &char &2.8 &26.2 &-1.3086 &2.9 &25.2 &-1.3019 \\
& &4k &2.8 &26.4 &-1.2933 &2.5 &26.6 &-1.2995 \\
& &32k &\textbf{3.0} &\textbf{28.3} &\textbf{-1.2445} &\textbf{3.1} &\textbf{27.5} &\textbf{-1.2623} \\  \cmidrule{2-9}
&\multirow{3}{*}{500k} &char &\textbf{12.9} &\textbf{45.7} &\textbf{0.0855} &11.8 &\textbf{43.4} &\textbf{-0.0212} \\
& &4k &11.1 &42.0 &-0.1612 &11.1 &41.8 &-0.1580 \\
& &32k &11.4 &42.3 &-0.0943 &\textbf{12.0} &42.5 &-0.0934 \\  \cmidrule{2-9}
&\multirow{3}{*}{5M} &char &17.3 &\textbf{50.7} &\textbf{0.6280} &\textbf{17.6} &\textbf{50.1} &0.6102 \\
& &4k &17.3 &49.8 &0.6140 &17.4 &49.8 &0.6045 \\
& &32k &\textbf{17.7} &49.9 &\textbf{0.6280} &17.5 &50.0 &\textbf{0.6409} \\ \midrule
\multirow{9}{*}{de} &\multirow{3}{*}{50k} &char &\textbf{5.7} &\textbf{35.4} &\textbf{-1.2272} &\textbf{5.0} &\textbf{33.0} &-1.2836 \\
& &4k &3.5 &31.5 &-1.3532 &3.2 &31.0 &-1.3571 \\
& &32k &4.8 &34.2 &-1.2328 &3.8 &32.9 &\textbf{-1.2819} \\  \cmidrule{2-9}
&\multirow{3}{*}{500k} &char &\textbf{18.9} &\textbf{51.1} &\textbf{0.3203} &\textbf{18.6} &\textbf{51.0} &\textbf{0.3155} \\
& &4k &17.1 &49.1 &0.1909 &16.6 &48.4 &0.1292 \\
& &32k &17.7 &48.8 &0.1595 &17.5 &49.0 &0.1624 \\  \cmidrule{2-9}
&\multirow{3}{*}{5M} &char &24.1 &\textbf{55.4} &0.6146 &24.1 &\textbf{54.9} &0.6007 \\
& &4k &24.6 &55.3 &0.6138 &24.1 &54.8 &0.6006 \\
& &32k &\textbf{24.8} &55.2 &\textbf{0.6178} &\textbf{24.3} &54.7 &\textbf{0.6055} \\ \midrule
\multirow{9}{*}{es} &\multirow{3}{*}{50k} &char &4.6 &32.8 &\textbf{-1.2302} &\textbf{4.5} &31.3 &\textbf{-1.2476} \\
& &4k &4.1 &30.7 &-1.2826 &3.3 &30.0 &-1.2983 \\
& &32k &\textbf{5.1} &\textbf{33.6} &-1.1571 &\textbf{4.5} &\textbf{32.6} &-1.1992 \\  \cmidrule{2-9}
&\multirow{3}{*}{500k} &char &\textbf{15.5} &\textbf{45.7} &\textbf{0.1277} &\textbf{14.8} &\textbf{45.6} &\textbf{0.0684} \\
& &4k &15.0 &44.6 &0.0258 &14.3 &43.8 &-0.0695 \\
& &32k &14.6 &44.1 &-0.0454 &\textbf{14.8} &44.1 &-0.0491 \\  \cmidrule{2-9}
&\multirow{3}{*}{5M} &char &\textbf{20.1} &\textbf{49.7} &\textbf{0.4917} &\textbf{19.8} &\textbf{49.1} &0.4679 \\
& &4k &19.3 &48.8 &0.4712 &19.6 &49.0 &0.4582 \\
& &32k &20.0 &48.9 &0.4670 &19.9 &49.0 &\textbf{0.4708} \\ \midrule
\multirow{6}{*}{hr} &\multirow{3}{*}{50k} &char &\textbf{10.3} &\textbf{42.3} &\textbf{-0.4010} &\textbf{9.5} &\textbf{40.4} &\textbf{-0.4877} \\
& &4k &5.7 &35.5 &-0.9234 &4.5 &33.3 &-1.0641 \\
& &32k &7.8 &37.9 &-0.7439 &6.7 &35.8 &-0.8185 \\  \cmidrule{2-9}
&\multirow{3}{*}{500k} &char &\textbf{19.3} &\textbf{51.6} &\textbf{0.6619} &\textbf{20.1} &\textbf{51.6} &\textbf{0.6795} \\
& &4k &18.0 &50.0 &0.5527 &18.6 &50.2 &0.5224 \\
& &32k &18.0 &49.6 &0.5050 &18.3 &49.6 &0.5208 \\
\bottomrule
\end{tabular}
\caption{Test set scores for deeper models (16 encoder layers, 6 decoder layers and 16 encoder layers, 16 decoder layers). Bold are the best results within the same training dataset and same model architecture.}\label{tab:deeper}
\end{table}

We compare BLEU, chrF and COMET scores for Transformer-base trained on different training dataset sizes and with different segmentations in all the language directions in Table \ref{tab:baseline} and the same results are plotted in Figure \ref{fig:sim_vs_vocab}. First and foremost, the character-level models provide the best results for the most similar language pair, Czech-Slovak (sk), across training data sizes and translation directions. For example, with a 50k dataset, the character-level model achieves a COMET score of $0.8834$ and $0.8429$ in Czech-to-Slovak and Slovak-to-Czech translations, respectively. The scores are better compared to those of 4k and 32k vocab models with the same training dataset. This trend continues with larger datasets; the character-level model outperforms in both the 500k and 5M datasets, although for the largest datasets, the results are very similar across vocabulary sizes.

 However, for the other language pairs, the results are mixed, and subword-level models often outperform character-level models, particularly with larger training dataset sizes. For instance, in Czech-to-Hungarian (hu) translations with a 5M dataset, the 32k vocab model achieves a COMET score of $0.6531$ which is better than the $0.6263$ score of the character-level model. The same pattern is observed in Czech-to-German (de) translations with the 32k vocab model outperforming the character-level model in the 5M dataset with a COMET score of $0.6275$ against $0.5955$. 
 
 For all the other languages (aside from Slovak), training on the 50k dataset fails to produce usable translation model at any vocabulary size, even for the second most similar language, Croatian. However, as we show in the next section, we can see the benefits of char-level translation of Czech-Croatian when finetuning charl-level model from subword-level model.
 
 The results are more favorable for subword-level models with increasing training set sizes, probably due to the sparsity of the longer subwords in smaller datasets which results in worse quality of the embeddings. We also see that generally, character-level models perform better in terms of chrF (char-level metric) than BLEU and COMET. For example, see Czech-to-Spanish, 5M dataset: character model has the best chrF score (although by a small margin), but the worst BLEU and COMET scores.
% \bibliography{\confname}
\subsection{Finetuning}
We took an alternative approach to training character-level models from scratch by fine-tuning the subword-level models. We only finetuned the models in the direction from Czech to the target language. Starting from the last checkpoint of the subword-level training, we switched the dataset to a character-split one. Since SentencePiece models include all the characters in their vocabularies, there was no need to adjust them. We proceeded with the same hyperparameters, including the optimizer parameters, after resetting the early-stopping counters.

We present the results in Tables \ref{tab:finetuned4} and \ref{tab:finetuned32} for models finetuned from 4k and 32k subword models, respectively. We see that in cases where training a char-level model from scratch didn't perform well compared to a subword-level one, finetuning from subword-level helps to attain the quality of the subword-level and even surpass it in some cases. For example, Czech-to-Croatian char level model without finetuning obtains COMET score of $-1.4055$, but after finetuning from 4k model, the score increases to $-0.2671$, which is also better than the $-1.0112$ of the 4k model alone.

Similar, although small increases compared to training from scratch can be seen across all the language pairs, with the exception of Czech-Slovak. For this pair, the translation quality of the character-level model trained from scratch is already much higher on the 50k and 500k datasets. Finetuning from either 32k or 4k models hurts the quality in this case, which could be expected.

 After the finetuning, the char-level Croatian model clearly outperforms both 4k and 32k subword models on the 50k dataset in all the metrics. As this did not occur with other, less similar languages, we hypothesize that language similarity is again an important factor in favor of character-level translation.

\subsection{Model size}
Previous work suggests that character-level processing in Transformers requires the use of deeper models to reach the same performance as subword-level processing. We present experiments with increasing depth of the model in Table \ref{tab:deeper}. All the models are trained in the direction Czech to target. The model sizes are described in Section \ref{sec:model}. We observe improvements in character-level translation compared to subword-level models of the same depth, but not compared to the \texttt{Transformer-base} models (the results are actually often worse than for the base model).  For instance, in German (de) target language with the 500k dataset, the character-level model using 16 encoder layers and 6 decoder layers yielded a COMET score of 0.3203. In contrast, the 4k and 32k vocab subword-level models achieved lower scores of 0.1909 and 0.1595, respectively. Similar patterns can be observed for other languages and datasets as well. However, the vanilla Transformer-base with 4k (Table \ref{tab:baseline}) obtained COMET of 0.3568, still outperforming even the deeper character-level model. The baseline models outperform the deeper models with 4k and 32k vocabularies, often by a large margin, while performance at char-level remains similar or only slightly worse (compare corresponding rows in Table \ref{tab:baseline} and Table \ref{tab:deeper}).

We hypothesize that the absence of improvements is caused by small dataset sizes and non-optimal hyperparameter choices. The results however suggest that deeper models are better suited for character-level translation, even though they mostly fail to outperform the shallower models in our setting. 

\section{Conclusions}
We trained standard Transformer models to translate between languages with different levels of similarity both on subword-segmented and character-segmented data.  We also varied the model depth and the training set size. We show that character-level models outperform subword-segmented models on the most closely related language pair (Czech-Slovak) as measured by automated MT quality metrics. Finetuning models trained with subword-level segmentation to character-level increases the performance in some cases. After finetuning, character-level models surpass the quality of subword-level models also for Czech-Croatian. Other, less similar language pairs reach similar preformances for both subword- and character-level models.

\section*{Acknowledgements}
This research was partially supported by grant 19-26934X  (NEUREM3)  of  the  Czech  Science Foundation and by the Charles University project GAUK No. 244523.

\bibliography{custom,anthology}

\begin{thebibliography}{}

\bibitem[Akhbardeh et~al., 2021]{akhbardeh-etal-2021-findings}
Akhbardeh, F., Arkhangorodsky, A., Biesialska, M., Bojar, O., Chatterjee, R.,
  Chaudhary, V., Costa-jussa, M.~R., Espa{\~n}a-Bonet, C., Fan, A., Federmann,
  C., Freitag, M., Graham, Y., Grundkiewicz, R., Haddow, B., Harter, L.,
  Heafield, K., Homan, C., Huck, M., Amponsah-Kaakyire, K., Kasai, J.,
  Khashabi, D., Knight, K., Kocmi, T., Koehn, P., Lourie, N., Monz, C.,
  Morishita, M., Nagata, M., Nagesh, A., Nakazawa, T., Negri, M., Pal, S.,
  Tapo, A.~A., Turchi, M., Vydrin, V., and Zampieri, M. (2021).
\newblock Findings of the 2021 conference on machine translation ({WMT}21).
\newblock In {\em Proceedings of the Sixth Conference on Machine Translation},
  pages 1--88, Online. Association for Computational Linguistics.

\bibitem[Banar et~al., 2021]{banar}
Banar, N., Daelemans, W., and Kestemont, M. (2021).
\newblock Character-level transformer-based neural machine translation.
\newblock In {\em Proceedings of the 4th International Conference on Natural
  Language Processing and Information Retrieval}, NLPIR 2020, page 149–156,
  New York, NY, USA. Association for Computing Machinery.

\bibitem[Ba{\~n}{\'o}n et~al., 2020]{banon-etal-2020-paracrawl}
Ba{\~n}{\'o}n, M., Chen, P., Haddow, B., Heafield, K., Hoang, H.,
  Espl{\`a}-Gomis, M., Forcada, M.~L., Kamran, A., Kirefu, F., Koehn, P.,
  Ortiz~Rojas, S., Pla~Sempere, L., Ram{\'\i}rez-S{\'a}nchez, G., Sarr{\'\i}as,
  E., Strelec, M., Thompson, B., Waites, W., Wiggins, D., and Zaragoza, J.
  (2020).
\newblock {P}ara{C}rawl: Web-scale acquisition of parallel corpora.
\newblock In {\em Proceedings of the 58th Annual Meeting of the Association for
  Computational Linguistics}, pages 4555--4567, Online. Association for
  Computational Linguistics.

\bibitem[Barrault et~al., 2019]{barrault-etal-2019-findings}
Barrault, L., Bojar, O., Costa-juss{\`a}, M.~R., Federmann, C., Fishel, M.,
  Graham, Y., Haddow, B., Huck, M., Koehn, P., Malmasi, S., Monz, C.,
  M{\"u}ller, M., Pal, S., Post, M., and Zampieri, M. (2019).
\newblock Findings of the 2019 conference on machine translation ({WMT}19).
\newblock In {\em Proceedings of the Fourth Conference on Machine Translation
  (Volume 2: Shared Task Papers, Day 1)}, pages 1--61, Florence, Italy.
  Association for Computational Linguistics.

\bibitem[Bella et~al., 2021]{bella}
Bella, G., Batsuren, K., and Giunchiglia, F. (2021).
\newblock A database and visualization of the similarity of contemporary
  lexicons.
\newblock In {\em Text, Speech, and Dialogue: 24th International Conference,
  TSD 2021, Olomouc, Czech Republic, September 6–9, 2021, Proceedings}, page
  95–104, Berlin, Heidelberg. Springer-Verlag.

\bibitem[Chung et~al., 2016]{chung-etal-2016-character}
Chung, J., Cho, K., and Bengio, Y. (2016).
\newblock A character-level decoder without explicit segmentation for neural
  machine translation.
\newblock In {\em Proceedings of the 54th Annual Meeting of the Association for
  Computational Linguistics (Volume 1: Long Papers)}, pages 1693--1703, Berlin,
  Germany. Association for Computational Linguistics.

\bibitem[Costa-juss{\`a} et~al., 2017]{costa-jussa-etal-2017-byte}
Costa-juss{\`a}, M.~R., Escolano, C., and Fonollosa, J. A.~R. (2017).
\newblock Byte-based neural machine translation.
\newblock In {\em Proceedings of the First Workshop on Subword and Character
  Level Models in {NLP}}, pages 154--158, Copenhagen, Denmark. Association for
  Computational Linguistics.

\bibitem[Gao et~al., 2020]{gao-etal-2020-character}
Gao, Y., Nikolov, N.~I., Hu, Y., and Hahnloser, R.~H. (2020).
\newblock Character-level translation with self-attention.
\newblock In {\em Proceedings of the 58th Annual Meeting of the Association for
  Computational Linguistics}, pages 1591--1604, Online. Association for
  Computational Linguistics.

\bibitem[Gupta et~al., 2019]{gupta2019characterbased}
Gupta, R., Besacier, L., Dymetman, M., and Gallé, M. (2019).
\newblock Character-based nmt with transformer.

\bibitem[Jon et~al., 2021]{jon-etal-2021-cuni}
Jon, J., Nov{\'a}k, M., Aires, J.~P., Varis, D., and Bojar, O. (2021).
\newblock {CUNI} systems for {WMT}21: Multilingual low-resource translation for
  {I}ndo-{E}uropean languages shared task.
\newblock In {\em Proceedings of the Sixth Conference on Machine Translation},
  pages 354--361, Online. Association for Computational Linguistics.

\bibitem[Junczys-Dowmunt et~al., 2018]{junczys-dowmunt-etal-2018-marian}
Junczys-Dowmunt, M., Grundkiewicz, R., Dwojak, T., Hoang, H., Heafield, K.,
  Neckermann, T., Seide, F., Germann, U., Aji, A.~F., Bogoychev, N., Martins,
  A. F.~T., and Birch, A. (2018).
\newblock {M}arian: Fast neural machine translation in {C}++.
\newblock In {\em Proceedings of {ACL} 2018, System Demonstrations}, pages
  116--121, Melbourne, Australia. Association for Computational Linguistics.

\bibitem[Kingma and Ba, 2017]{kingma2017adam}
Kingma, D.~P. and Ba, J. (2017).
\newblock Adam: A method for stochastic optimization.

\bibitem[Lee et~al., 2017]{lee-etal-2017-fully}
Lee, J., Cho, K., and Hofmann, T. (2017).
\newblock Fully character-level neural machine translation without explicit
  segmentation.
\newblock {\em Transactions of the Association for Computational Linguistics},
  5:365--378.

\bibitem[Li et~al., 2021]{li-etal-2021-char}
Li, J., Shen, Y., Huang, S., Dai, X., and Chen, J. (2021).
\newblock When is char better than subword: A systematic study of segmentation
  algorithms for neural machine translation.
\newblock In {\em Proceedings of the 59th Annual Meeting of the Association for
  Computational Linguistics and the 11th International Joint Conference on
  Natural Language Processing (Volume 2: Short Papers)}, pages 543--549,
  Online. Association for Computational Linguistics.

\bibitem[Libovick{\'y} and Fraser, 2020]{libovicky-fraser-2020-towards}
Libovick{\'y}, J. and Fraser, A. (2020).
\newblock Towards reasonably-sized character-level transformer {NMT} by
  finetuning subword systems.
\newblock In {\em Proceedings of the 2020 Conference on Empirical Methods in
  Natural Language Processing (EMNLP)}, pages 2572--2579, Online. Association
  for Computational Linguistics.

\bibitem[Libovick{\'y} et~al., 2022]{libovicky-etal-2022-dont}
Libovick{\'y}, J., Schmid, H., and Fraser, A. (2022).
\newblock Why don{'}t people use character-level machine translation?
\newblock In {\em Findings of the Association for Computational Linguistics:
  ACL 2022}, pages 2470--2485, Dublin, Ireland. Association for Computational
  Linguistics.

\bibitem[Ngo et~al., 2019]{ngo-etal-2019-Transformer}
Ngo, T.-V., Ha, T.-L., Nguyen, P.-T., and Nguyen, L.-M. (2019).
\newblock How transformer revitalizes character-based neural machine
  translation: An investigation on {J}apanese-{V}ietnamese translation systems.
\newblock In {\em Proceedings of the 16th International Conference on Spoken
  Language Translation}, Hong Kong. Association for Computational Linguistics.

\bibitem[Popovi{\'c}, 2015]{popovic-2015-chrf}
Popovi{\'c}, M. (2015).
\newblock chr{F}: character n-gram {F}-score for automatic {MT} evaluation.
\newblock In {\em Proceedings of the Tenth Workshop on Statistical Machine
  Translation}, pages 392--395, Lisbon, Portugal. Association for Computational
  Linguistics.

\bibitem[Post, 2018]{post-2018-call}
Post, M. (2018).
\newblock A call for clarity in reporting {BLEU} scores.
\newblock In {\em Proceedings of the Third Conference on Machine Translation:
  Research Papers}, pages 186--191, Brussels, Belgium. Association for
  Computational Linguistics.

\bibitem[Rei et~al., 2020]{rei-etal-2020-comet}
Rei, R., Stewart, C., Farinha, A.~C., and Lavie, A. (2020).
\newblock {COMET}: A neural framework for {MT} evaluation.
\newblock In {\em Proceedings of the 2020 Conference on Empirical Methods in
  Natural Language Processing (EMNLP)}, pages 2685--2702, Online. Association
  for Computational Linguistics.

\bibitem[Scherrer et~al., 2019]{scherrer-etal-2019-university}
Scherrer, Y., V{\'a}zquez, R., and Virpioja, S. (2019).
\newblock The {U}niversity of {H}elsinki submissions to the {WMT}19 similar
  language translation task.
\newblock In {\em Proceedings of the Fourth Conference on Machine Translation
  (Volume 3: Shared Task Papers, Day 2)}, pages 236--244, Florence, Italy.
  Association for Computational Linguistics.

\bibitem[Team et~al., 2022]{nllb2022}
Team, N., Costa-jussà, M.~R., Cross, J., Çelebi, O., Elbayad, M., Heafield,
  K., Heffernan, K., Kalbassi, E., Lam, J., Licht, D., Maillard, J., Sun, A.,
  Wang, S., Wenzek, G., Youngblood, A., Akula, B., Barrault, L., Gonzalez,
  G.~M., Hansanti, P., Hoffman, J., Jarrett, S., Sadagopan, K.~R., Rowe, D.,
  Spruit, S., Tran, C., Andrews, P., Ayan, N.~F., Bhosale, S., Edunov, S., Fan,
  A., Gao, C., Goswami, V., Guzmán, F., Koehn, P., Mourachko, A., Ropers, C.,
  Saleem, S., Schwenk, H., and Wang, J. (2022).
\newblock No language left behind: Scaling human-centered machine translation.

\bibitem[Vaswani et~al., 2017]{vaswani2017attention}
Vaswani, A., Shazeer, N., Parmar, N., Uszkoreit, J., Jones, L., Gomez, A.~N.,
  Kaiser, L., and Polosukhin, I. (2017).
\newblock Attention is all you need.

\end{thebibliography}
\bibliographystyle{apalike}
\end{document}